\documentclass{article}

\PassOptionsToPackage{numbers, compress}{natbib}
\usepackage[preprint]{neurips_2026}
\usepackage[utf8]{inputenc}
\usepackage[T1]{fontenc}
\usepackage{hyperref}
\usepackage{url}
\usepackage{graphicx}
\usepackage{booktabs}
\usepackage{wrapfig}
\usepackage{booktabs}
\usepackage{amsmath}
\usepackage{amsfonts}
\usepackage{nicefrac}
\usepackage{microtype}
\usepackage{xcolor}
\newcommand{\Description}[1]{}

\usepackage{multirow}
\usepackage{amsthm}

\newtheorem{proposition}{Proposition}

\begin{document}

\title{Asymmetric Invertible Threat: Learning Reversible Privacy Defense for Face Recognition}

\author{%
  Jiabei Zhang\textsuperscript{1} \quad
  Ziyuan Yang\textsuperscript{2} \quad
  Andrew Beng Jin Teoh\textsuperscript{3} \quad
  Yi Zhang\textsuperscript{1} \\
  \\
  \textsuperscript{1}School of Cyber Science and Engineering, Sichuan University \\
  \textsuperscript{2}Lee Kong Chian School of Medicine, Nanyang Technological University \\
  \textsuperscript{3}School of Electrical and Electronic Engineering, Yonsei University \\
  \\
  \texttt{2023141530097@stu.scu.edu.cn} \quad
  \texttt{cziyuanyang@gmail.com} \\
  \texttt{bjteoh@yonsei.ac.kr} \quad
  \texttt{yzhang@scu.edu.cn}
}

\maketitle


\begin{abstract}
Face Recognition systems are widely deployed in real-world applications, but they also raise privacy concerns due to unauthorized collection and misuse of facial data. Existing adversarial privacy protection methods rely on input-space perturbations to obfuscate identity information, yet their protection can degrade when adversaries learn restoration or purification mappings that partially invert the transformation. We study this setting as an asymmetric adversarial attack, in which reverse manipulation becomes feasible because existing defense paradigms do not control reversibility. To address this problem, we propose Asymmetric Reversible Face Protection (ARFP), a restoration-aware extension of personalized face cloaking that integrates privacy protection, keyed recovery, and tamper indication in a single framework. ARFP consists of three components: Key-Conditioned Manifold Binding, which ties the protection transformation to a user-provided key; Adversarial Restoration-Aware Training, which introduces a surrogate restoration adversary during training to improve robustness against evaluated inverse purification attacks; and Authorized Reversible Restoration, which supports recovery with the correct key while providing nonce-based tamper indication. Extensive experiments under the threat models considered in this work show that ARFP improves resistance to the evaluated restoration attacks while preserving authorized recovery utility. These results provide empirical evidence of key-sensitive recovery behavior and tamper awareness in the tested settings.
\end{abstract}

\section{Introduction}

The widespread deployment of Face Recognition (FR) systems~\cite{schroff2015facenet,deng2019arcface,wang2018cosface,meng2021magface} has transformed digital identity verification, supporting applications ranging from secure authentication to smart surveillance. 
However, it also raises substantial privacy concerns. Facial images shared online can be scraped to construct large-scale databases without user consent~\cite{shan2020fawkes,cherepanova2021lowkey}, enabling misuse such as covert tracking, identity theft, and behavioral profiling~\cite{sharif2016accessorize,oh2017adversarial}.

To mitigate such risks, adversarial privacy protection has emerged as a practical defense direction~\cite{shan2020fawkes,cherepanova2021lowkey}. Methods such asFawkes~\cite{shan2020fawkes}, LowKey~\cite{cherepanova2021lowkey}, and Chameleon~\cite{chow2024chameleon} perturb the input image to suppress unauthorized FR while preserving visual utility. Chameleon~\cite{chow2024chameleon}, in particular, introduces personalized privacy masks to improve efficiency and transferability. Despite these advances, existing methods primarily operate as input-space cloaks and do not explicitly control whether the applied transformation can later be approximated, neutralized, or partially reversed~\cite{goodfellow2014explaining,huang2021unlearnable,yin2021adv}.

Despite their practical effectiveness, these approaches rely on a one-sided protection assumption: once a perturbation or privacy mask is applied, the resulting image is implicitly treated as protected. This overlooks a fundamental asymmetry between intended access control and the model's actual capabilities. The forward transformation is meant to restrict identity inference, but no corresponding mechanism constrains reverse manipulation. Consequently, once restoration becomes feasible, it is not reserved for authorized users and may also be exploited by an adversary.

\begin{wrapfigure}{R}{0.48\textwidth}
    \centering
    \vspace{-0.5em}
    \includegraphics[width=\linewidth]{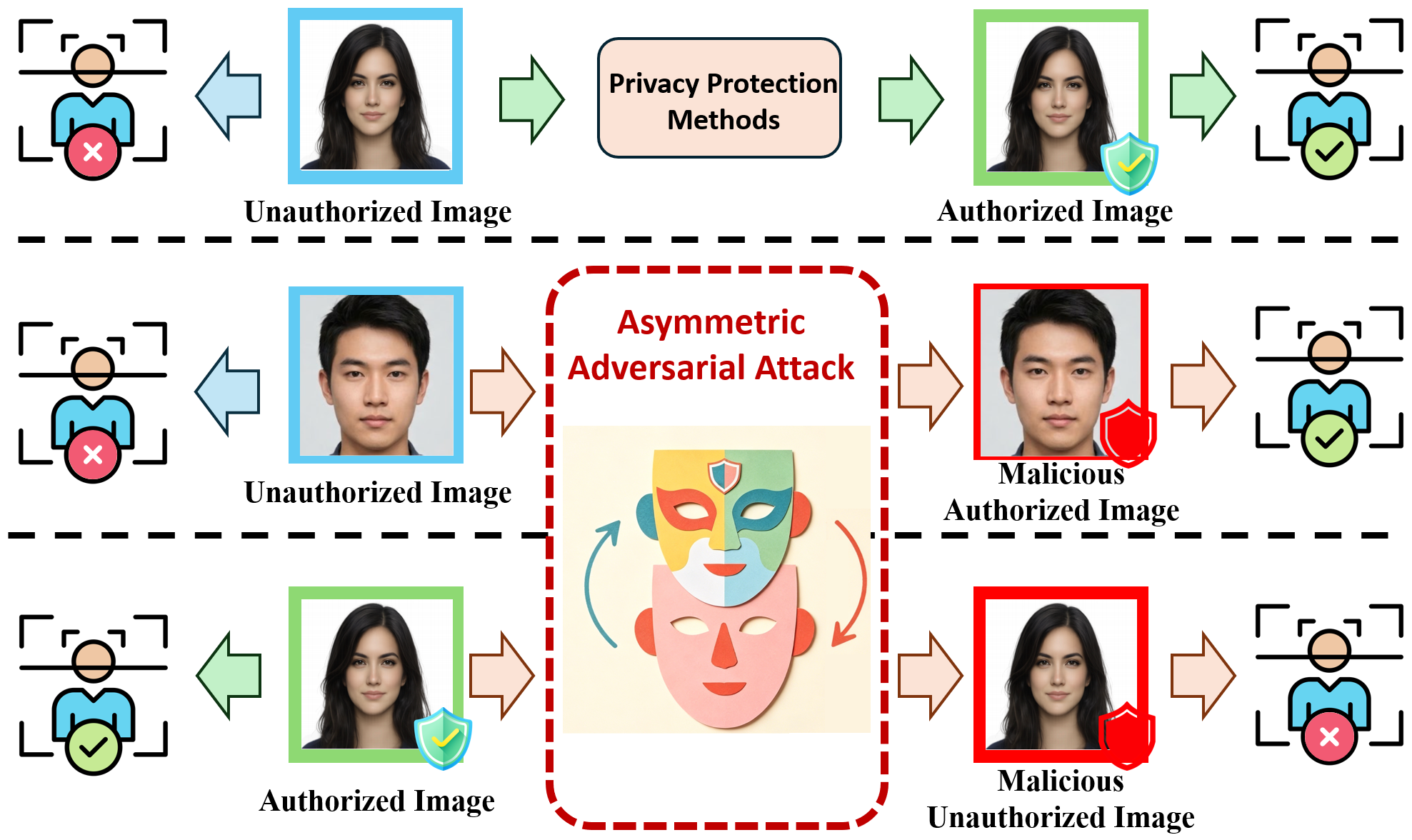}
    \vspace{-20pt}
    \caption{Illustration of the asymmetric adversarial attack.}
    \label{fig:attack_demo}
    \vspace{-1.5em}
\end{wrapfigure}

We study this vulnerability as an \textbf{\textit{asymmetric adversarial attack}}, in which an adversary learns or approximates an inverse mapping to reconstruct or weaken a privacy-preserving transformation without privileged access. Figure~\ref{fig:attack_demo} shows this setting. The vulnerability arises because existing cloaking methods remain in the same model space used for training and evaluation. If the protection transformation can be approximated there, reverse manipulation can also be learned. The practical question is not only whether a cloak disrupts recognition, but whether reversal is restricted to authorized parties.

Face privacy protection is therefore an \textit{asymmetric, bidirectional problem}: a forward mapping generates a protected image for public release, whereas the reverse mapping should be available only under authorized conditions. This motivates key-conditioned protection and direct robustness training against restoration attacks.

To address this problem, we propose \textit{\textbf{Asymmetric Reversible Face Protection (ARFP)}}, which is best viewed as a restoration-aware extension of personalized face cloaking. ARFP integrates three modules: Key-Conditioned Manifold Binding~(KMB), Adversarial Restoration-Aware Training~(ARAT), and Authorized Reversible Restoration~(ARR). KMB embeds a user-specific secret key into the mask generation process via affine modulation, producing a key-dependent transformation manifold. ARAT introduces a surrogate restoration network during training and optimizes the protector against this explicit inverse adversary. ARR supports key-guided recovery for authorized users while extracting a nonce as a fragile tamper-indication signal. Among these components, the central contribution is the restoration-aware training loop, which improves robustness to restoration attempts within the same protection pipeline. ARFP is model-agnostic and can be integrated with existing FR systems without retraining the recognizer.

The main contributions of this work are summarized as follows:
\begin{itemize}
    \item We study restoration-based attacks against adversarial face cloaking in the asymmetric adversarial attack setting, showing that exposure of reversibility can weaken existing input-space protections.
    \item We present ARFP as a restoration-aware extension of personalized face cloaking. Its central technical contribution is surrogate-restorer training, which improves robustness to restoration attempts within a key-conditioned protection pipeline.
    \item We integrate key-conditioned protection, authorized recovery, and nonce-based fragile-tamper indication into a single framework.
\end{itemize}

\section{Related Work}
Deep learning has substantially advanced FR performance~\cite{schroff2015facenet,deng2019arcface}.
However, FR systems remain vulnerable to adversarial manipulation~\cite{sharif2016accessorize,oh2017adversarial}. Early digital cloaking methods, such as Fawkes~\cite{shan2020fawkes}, generate identity-shifting perturbations but require costly optimization and exhibit limited robustness to common image transformations. LowKey~\cite{cherepanova2021lowkey} improves black-box transferability through ensemble-based attacks, but still incurs substantial computation and may introduce visible artifacts. Physical-world methods, including Adv-Makeup~\cite{yin2021adv} and AdvHat~\cite{komkov2021advhat}, target patch-based attacks and are less suited to protecting digital images. Chameleon~\cite{chow2024chameleon} further improves efficiency and transferability by generating personalized privacy masks with a conditional generative model.

Within this line of work, ARFP is more appropriately viewed as a restoration-aware extension of personalized face cloaking in the digital setting, rather than as a general solution to recoverable or verifiable privacy protection. The central distinction is the incorporation of a surrogate restoration adversary during training so that the learned protection remains more robust to the evaluated inverse or purification attacks, while still supporting keyed recovery and tamper indication.

\section{Security Analysis and Empirical Insights}
\subsection{Threat Model}
\noindent \textbf{\textit{Adversary.}} In our threat model, the adversary can access protected facial images as well as publicly available face images from the web, but has no access to the user’s secret key or the internal parameters of the protection mechanism. The adversary does not require knowledge of, or access to, the used FR model. Under such a setting, the adversary can readily collect large-scale facial data and perform identity inference without additional system-specific information.

\noindent \textbf{\textit{Defender.}} The defender operates without any special privileges over the FR system and does not modify or access its internal parameters. Under this setting, we identify the following defense goals:

\textbf{1. Rejection of Unauthorized Inputs.} Unauthorized face images should be rejected by the protected FR systems to reduce identity leakage from raw inputs.

\textbf{2. Visual Fidelity Preservation.} The protection process should preserve visual quality and usability for legitimate images sharing.

\textbf{3. Identity Obfuscation.} Protected images should suppress identity information cues against adversarial FR models while maintaining the intended asymmetric protection setting.

\textbf{4. Authorized Recovery.} The mechanism should support the recovery of identity-consistent images under the correct credential.

Existing methods mainly address the first two objectives. By contrast, our formulation is designed to address all four within a unified framework, with particular emphasis on the risk posed by inversion-oriented attacks under asymmetric access conditions.


\subsection{Empirical Motivation}

\begin{wrapfigure}{R}{0.48\textwidth}
    \centering
    \vspace{-1.5em}
    \includegraphics[width=\linewidth]{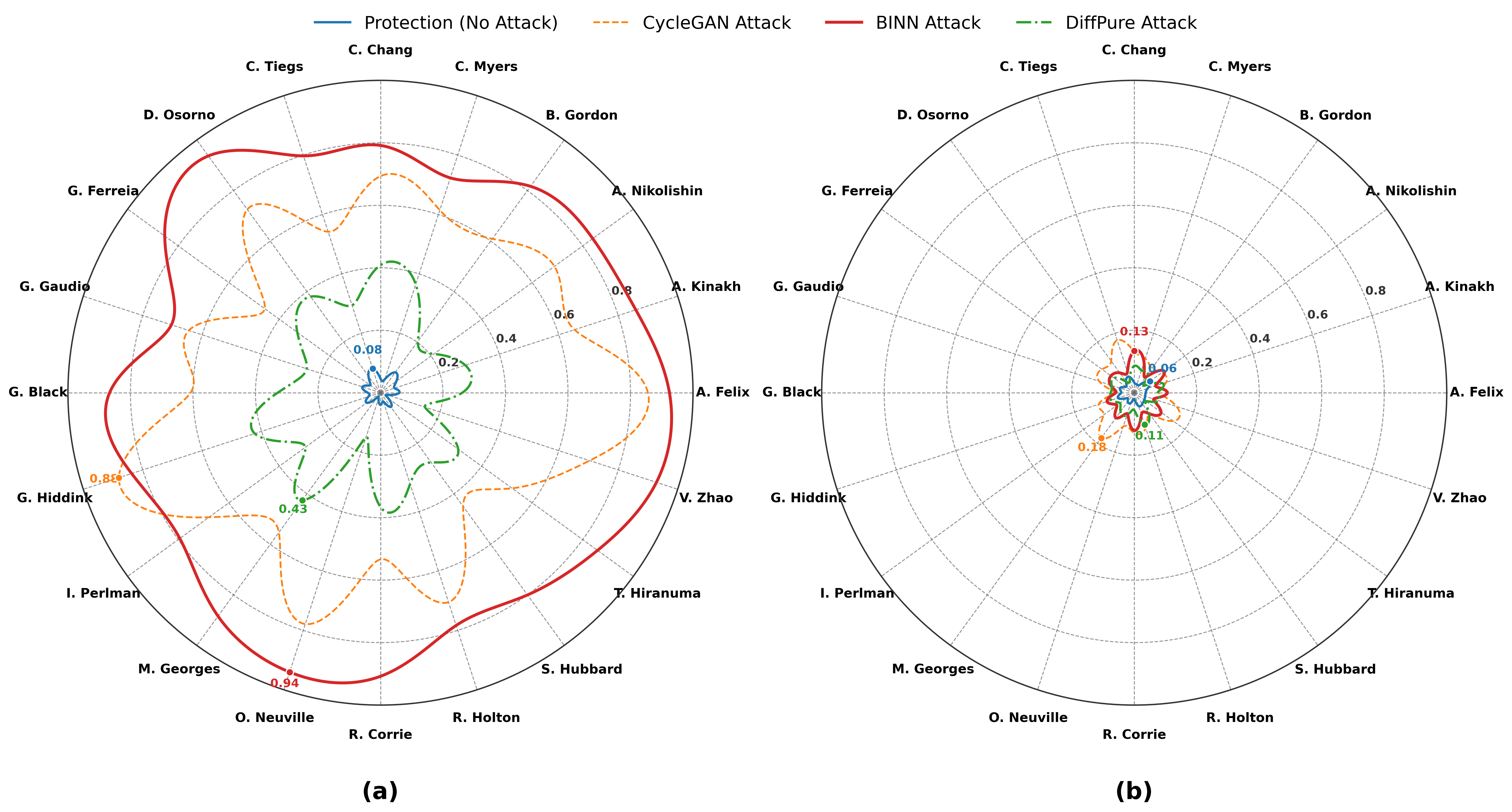}
    \vspace{-15pt}
    \caption{Feature Similarity under attacks (Lower is better). (a) Baseline fails against attacks. (b) ARFP maintains robust protection.}
    \vspace{-15pt}
    \label{fig:motivation_matrix}
\end{wrapfigure}

We conduct preliminary experiments to motivate our study by comparing representative prior work, Chameleon~\cite{chow2024chameleon},with ARFP under inversion-oriented transformations. We use several domain-translation models to simulate restoration or inversion attempts. As shown in Figure~\ref{fig:motivation_matrix}(a), Chameleon effectively rejects conventional recognition when such transformations are absent. However, under domain-transferred inputs, its protection degrades substantially for most evaluated inversion methods. In particular, unsupervised transformation models such as CycleGAN~\cite{zhu2017unpaired} and DiffPure~\cite{nie2022diffusion} reduce the protection effect, while the supervised domain-transfer method Adaptive Reversal causes more pronounced degradation. These results suggest that, for the transformation families evaluated here, static obfuscation may not remain stable under distribution-altering but label-consistent mappings.

In contrast, Figure~\ref{fig:motivation_matrix} (b) shows that ARFP maintains lower identity similarity under the same evaluated transformations. Its key-conditioned design appears more resistant to these inversion-oriented attacks in our empirical setting while preserving the asymmetric access structure intended by the framework. These preliminary results motivate the need for an asymmetric defense paradigm, though they should be interpreted within the scope of the evaluated attack models.

\section{Method}

\begin{figure*}[t]
    \centering
    \includegraphics[width=\linewidth]{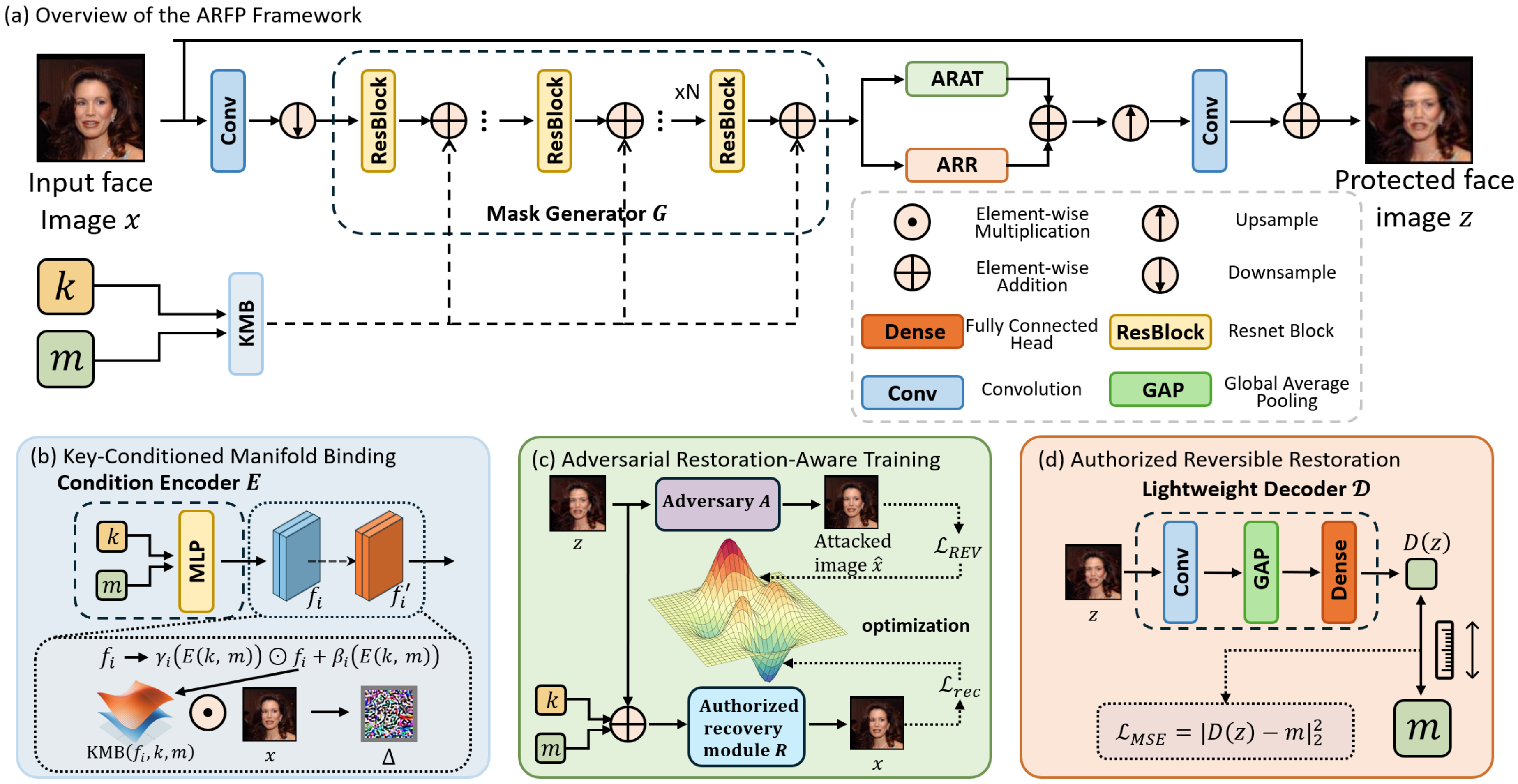}
    \Description{Diagram.}
    \caption{Overview of the proposed Asymmetric Reversible Face Protection (ARFP) framework. Our framework comprises three components: (b) KMB for key-conditioned feature modulation, (c) ARAT for robustness against restoration attacks, and (d) ARR for authorized reversible recovery.}
    \label{fig:framework}
    \vspace{-20pt}
\end{figure*}

\subsection{Overview of the ARFP Framework}
To preserve privacy and against inversion-oriented attacks, we propose the Asymmetric Reversible Face Protection (ARFP) framework, as shown in Figure~\ref{fig:framework}. ARFP formulates face protection as a key-conditioned generative process with three coupled components. Key-Conditioned Manifold Binding (KMB) injects a user-specific key into the protection mapping so the generated perturbation depends on both image content and credential. Adversarial Restoration-Aware Training (ARAT) introduces a surrogate restoration adversary during optimization to improve resistance to inversion attempts. Authorized Reversible Restoration (ARR) enables credential-conditioned recovery and provides an auxiliary tamper signal based on a nonce. Together, these components define a protection pipeline with keyed conditioning, restoration-aware training, and recovery support under asymmetric access conditions.

\subsection{Key-Conditioned Manifold Binding}
To condition protection on user credentials, we introduce the KMB module. The conditional mask generator $G$ is based on ResNet and receives an input image $x$, a secret key $k$, and a stochastic nonce $m$. Rather than using a fixed perturbation pattern, KMB modulates intermediate generator features with a condition embedding derived from $(k,m)$, so that the resulting protection depends jointly on image content and the provided credential.

Let $x \in \mathbb{R}^{3 \times H \times W}$ be the original image, and let $f_i \in \mathbb{R}^{C_i \times H_i \times W_i}$ denote the feature map at the $i$-th layer of the generator. We introduce a condition encoder $E$ implemented as a multi-layer perceptron, which maps the key--nonce pair $(k,m)$ to a latent embedding $e = E(k,m)$. KMB then recalibrates channel responses through learned affine modulation:
\begin{equation}
    \text{KMB}(f_i, k, m) = \gamma_i(E(k, m)) \odot f_i + \beta_i(E(k, m)),
\end{equation}
where $\gamma_i(E(k,m)), \beta_i(E(k,m)) \in \mathbb{R}^{C_i}$ denote channel-wise scale and shift parameters that are broadcast over the spatial dimensions. The final protective perturbation $\Delta(x, k, m)$ is produced by aggregating these modulated features across the generator, and the protected image is defined as:

\begin{equation}
    z = G(x, k, m) = x + \alpha \cdot \Delta(x, k, m),
\end{equation}
where $\alpha$ controls the perturbation magnitude.

Under this construction, $\Delta$ is not a fixed additive mask but an input- and key-dependent transformation induced by the modulated generator features. In this sense, KMB makes protection mapping conditional on the supplied credential, consistent with the empirical key-sensitivity analysis presented later, rather than implying a general guarantee of irreversibility.


\subsection{Adversarial Restoration-Aware Training}
Standard adversarial training typically optimizes against a classifier. By contrast, ARAT formulates face privacy protection as a minimax optimization problem against an explicit inversion adversary. We write the objective as:
\begin{equation}
    \min_{G, R} \max_{A} \mathbb{E}_{x, k, m} \Big[ \| x - R(G(x, k, m), k, m) \|_1 - \lambda_{rev} \mathcal{L}_{REV}(x, A(G(x, k, m))) \Big],
\end{equation}
where $A$ denotes a surrogate restoration adversary network~\cite{dinh2014nice}, and $R$ is the authorized recovery module. For convenience, we denote the authorized reconstruction term by:
\begin{equation}
    \mathcal{L}_{rec}(x,\hat{x}) = \|x - \hat{x}\|_1,
\end{equation}
with $\hat{x} = R(G(x,k,m),k,m)$. The adversarial restoration loss is defined as:
\begin{equation}
    \mathcal{L}_{REV}(x, \tilde{x}) = \|x - \tilde{x}\|_1 + \lambda_{l2} \|x - \tilde{x}\|_2^2,
\end{equation}
where $\tilde{x} = A(G(x,k,m))$ and $\lambda_{l2}$ controls the relative contribution of the $\ell_2$ term. The coefficient $\lambda_{rev}$ balances recovery fidelity and restoration resistance.

Under this formulation, $A$ is optimized to minimize $\mathcal{L}_{REV}$ and thereby approximate an inverse mapping from protected images $z$ to clean images $x$. In contrast, $G$ and $R$ are optimized so that authorized recovery remains feasible while the adversary's restoration becomes more difficult. In each training cycle, we alternate these updates: the surrogate adversary minimizes $\mathcal{L}_{REV}$ with the protector fixed, and the protector-side modules are then updated with the adversary fixed. This makes the optimization schedule explicit and removes ambiguity in the compact minimax notation.

The role of ARAT is therefore restoration-aware rather than purely classifier-aware. Its objective is to train the protection mapping against the specific restoration pressure induced by $A$, while preserving recoverability under the provided key.

\subsection{Authorized Reversible Restoration}
We next introduce the ARR module, which provides credential-conditioned recovery together with a lightweight tamper-indication signal. A nonce $m$ is embedded during generation as an auxiliary conditioning variable. A lightweight decoder $D$ is trained to recover this nonce from the protected image $z$:
\begin{equation}
\mathcal{L}_{MSE} = \| D(z) - m \|_2^2.
\end{equation}

At inference time, the same decoder can be applied to $z$ and, when needed, to the recovered image $\hat{x}$ to check whether the embedded signal remains consistent with the original nonce.

Consistency between the extracted nonce $\hat{m}$ and the original $m$ serves as a practical indicator of significant modification. This branch is intended as fragile tamper indication or provenance awareness, not as cryptographic proof of ownership.


The total training objective is:
\begin{equation}
      \mathcal{L}_{Total} = \lambda_{rec}\mathcal{L}_{rec} + \lambda_{id}\mathcal{L}_{id} + \lambda_{MSE}\mathcal{L}_{MSE} - \lambda_{rev}\mathcal{L}_{REV},
\end{equation}
where $\mathcal{L}_{id}$ is the identity loss, implemented by maximizing cosine distance in FR feature space to suppress identity matching. The hyperparameters $\lambda_{rec}$, $\lambda_{id}$, and $\lambda_{MSE}$ control the relative weights of recovery fidelity, identity obfuscation, and nonce consistency, respectively. The term $-\lambda_{rev}\mathcal{L}_{REV}$ couples the protector to the surrogate adversary by penalizing solutions that are easy to restore.

In practice, once $A$ is updated to reduce $\mathcal{L}_{REV}$, the protector-side modules are updated to minimize $\mathcal{L}_{Total}$ with $A$ fixed. This alternating procedure clarifies how ARR interacts with ARAT during optimization.


\subsection{Information-Theoretic Interpretation}

This section provides a supportive information-theoretic interpretation of feature leakage under the proposed key-conditioned transformation. Its purpose is to formalize one aspect of the protection mechanism at the level of identity-feature estimation, rather than to provide a formal cryptographic security guarantee. Specifically, let \(G: \mathcal{X} \times \mathcal{K} \rightarrow \mathcal{Z}\) denote the protection mapping, where \(x \in \mathcal{X}\) is an input face image and \(k \in \mathcal{K}\) is a secret key sampled independently from a high-entropy key space. The protected output is \(z = G(x,k)\). Let \(\phi: \mathcal{X} \rightarrow \mathbb{R}^{d}\) be a fixed face-recognition model, and let \(h = \phi(x)\) denote the corresponding identity embedding.

\begin{proposition}[Feature-leakage interpretation under protected observation]
\label{prop:feature_leakage}
Consider an adversary that estimates the identity embedding \(h\) using only the protected observation \(z\) under squared error. Then the optimal estimator is given by the conditional expectation of \(h\) given \(z\), and the corresponding minimum estimation error is governed by the conditional uncertainty of \(h\) after observing \(z\). Consequently, when the mutual information between \(h\) and \(z\) is reduced, the remaining uncertainty about \(h\) given \(z\) increases, which is consistent with greater difficulty in recovering identity-related features from the protected output.
\end{proposition}

\begin{wrapfigure}{R}{0.48\textwidth}
    \centering
    \vspace{-30pt}
    \includegraphics[width=0.45\textwidth]{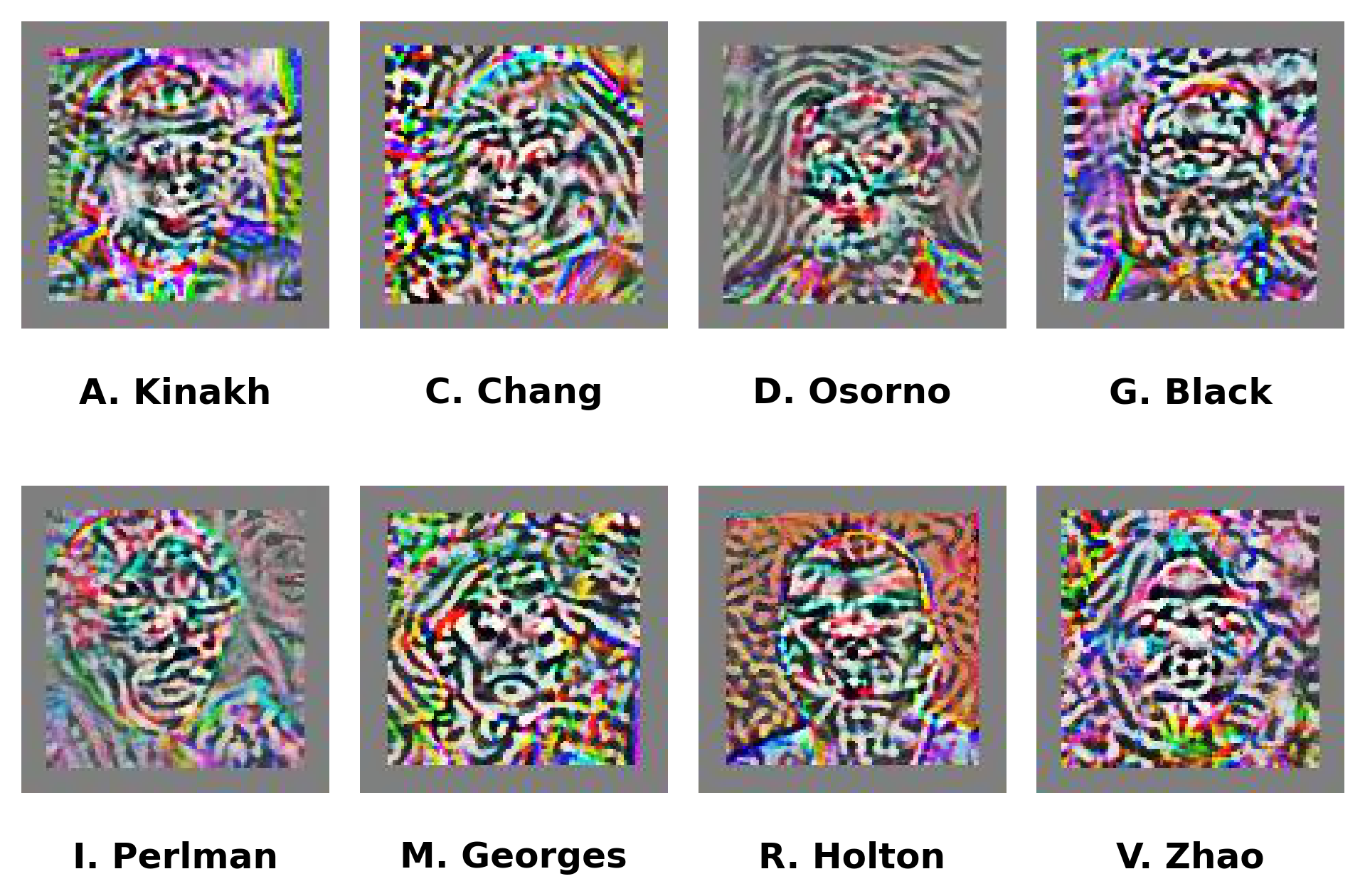}
    \Description{Diagram.}
    \caption{Visualization of rescaled protection masks. ARFP concentrates identity-specific perturbations on critical semantic regions.}
    \vspace{-20pt}
    \label{fig:mask_viz}
\end{wrapfigure}

Proposition~\ref{prop:feature_leakage} should be interpreted as a qualitative information-theoretic statement rather than as a closed-form security bound. Its role is to support the design intuition behind key-conditioned protection: if the transformation suppresses the information about \(h\) that remains observable through \(z\), then recovering identity-related features from the protected output becomes harder on average. This links the empirical objective of reducing feature leakage to a mathematically interpretable quantity, but it does not by itself imply a general guarantee of inversion resistance, irreversibility, or cryptographic security. A short supporting derivation is provided in Appendix~\ref{sec:appendix_proof}.

\section{Experiments}
\subsection{Experimental Setup}
\noindent\textbf{Datasets.} We evaluate our framework on two benchmark datasets widely used in face privacy research: LFW~\cite{huang2007labeled} and FaceScrub~\cite{ng2014data}.Detailed preprocessing operations are provided in Appendix~\ref{sec:appendix_exp}.

\noindent\textbf{Face Recognition Models.} To evaluate robustness and transferability, we use a diverse collection of publicly available pretrained FR models, including five architectures trained on four public datasets. Details are provided in Appendix~\ref{sec:appendix_exp}.

\begin{wrapfigure}{R}{0.48\textwidth}
    \vspace{-30pt}
    \centering
    \includegraphics[width=0.45\textwidth]{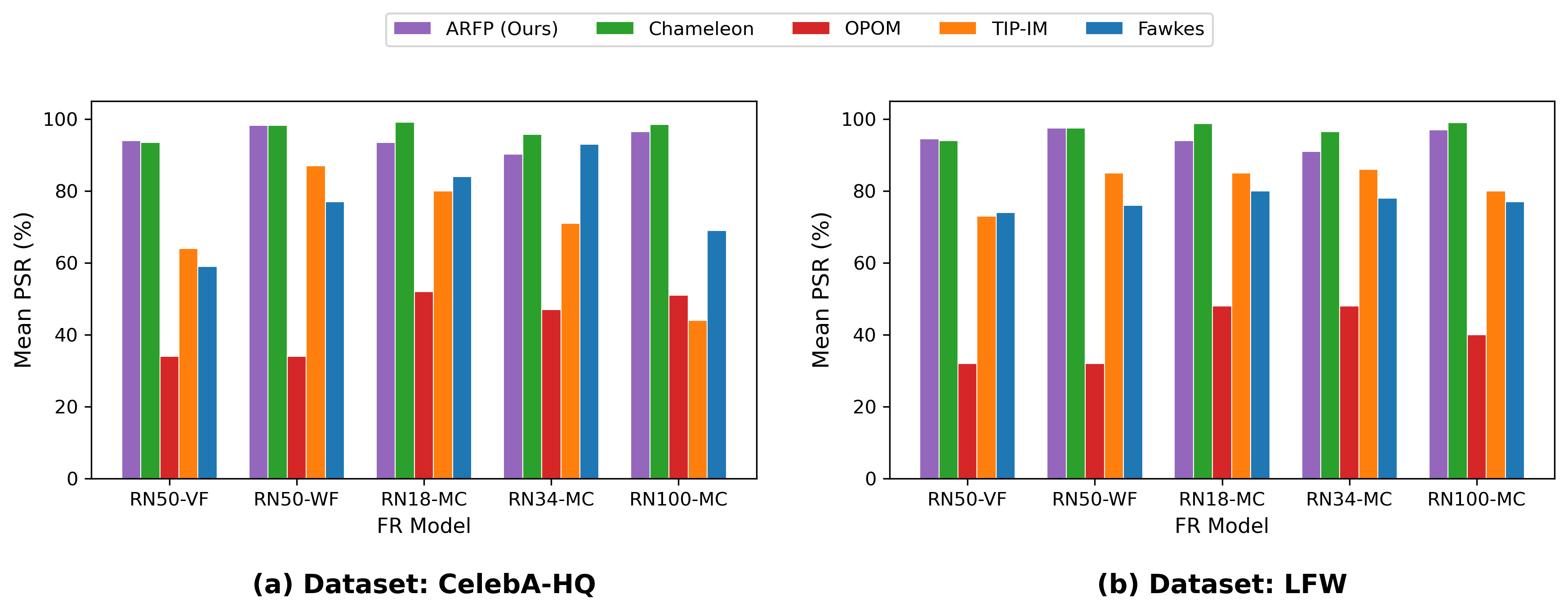}
    \Description{Diagram.}
    \vspace{-10pt}
    \caption{Protection effectiveness on LFW against unseen FR models.}
    \vspace{-15pt}
    \label{fig:comparison_chart}
  \end{wrapfigure}

\noindent\textbf{Baselines.} We compare ARFP against representative face-cloaking baselines under the relevant evaluation settings. For the unseen-model LFW comparison in Fig.~\ref{fig:comparison_chart}, we report methods with directly comparable results under that protocol: Chameleon~\cite{chow2024chameleon}, OPOM~\cite{zhong2022opom}, and TIP-IM~\cite{Yang_2021_ICCV}. For restoration-attack and image-quality evaluations, we additionally compare with Fawkes (High mode)~\cite{shan2020fawkes} and Chameleon~\cite{chow2024chameleon}. All baselines are run using their official implementations and recommended configurations.

\noindent\textbf{Implementation Details.} Comprehensive network and training hyperparameters are provided in Appendix~\ref{sec:appendix_exp}.

\noindent\textbf{Evaluation Metric.} Following prior work~\cite{chow2024chameleon}, we adopt the Protection Success Rate (PSR) to evaluate performance against unauthorized FR. PSR is defined as the complement of FR accuracy on the protected dataset:
\begin{equation}
    \text{PSR} = 100\% - \text{FR Accuracy}.
\end{equation}
A higher PSR indicates more effective prevention of matching to the ground-truth identity.

\subsection{Protection Performance}
To qualitatively assess the generated protection mechanisms, we visualize the learned masks and the protected images they produce. Unlike fixed or identity-agnostic masks, our approach produces adaptive perturbations for different faces, enabling more personalized protection.

Figure~\ref{fig:mask_viz} illustrates the learned masks for selected identities. The masks exhibit distinct identity-dependent structures driven by the KMB module, suggesting that the protection adapts to subject-specific facial content.

\begin{wrapfigure}{L}{0\textwidth}
    \vspace{-40pt}
    \includegraphics[width=0.45\textwidth]{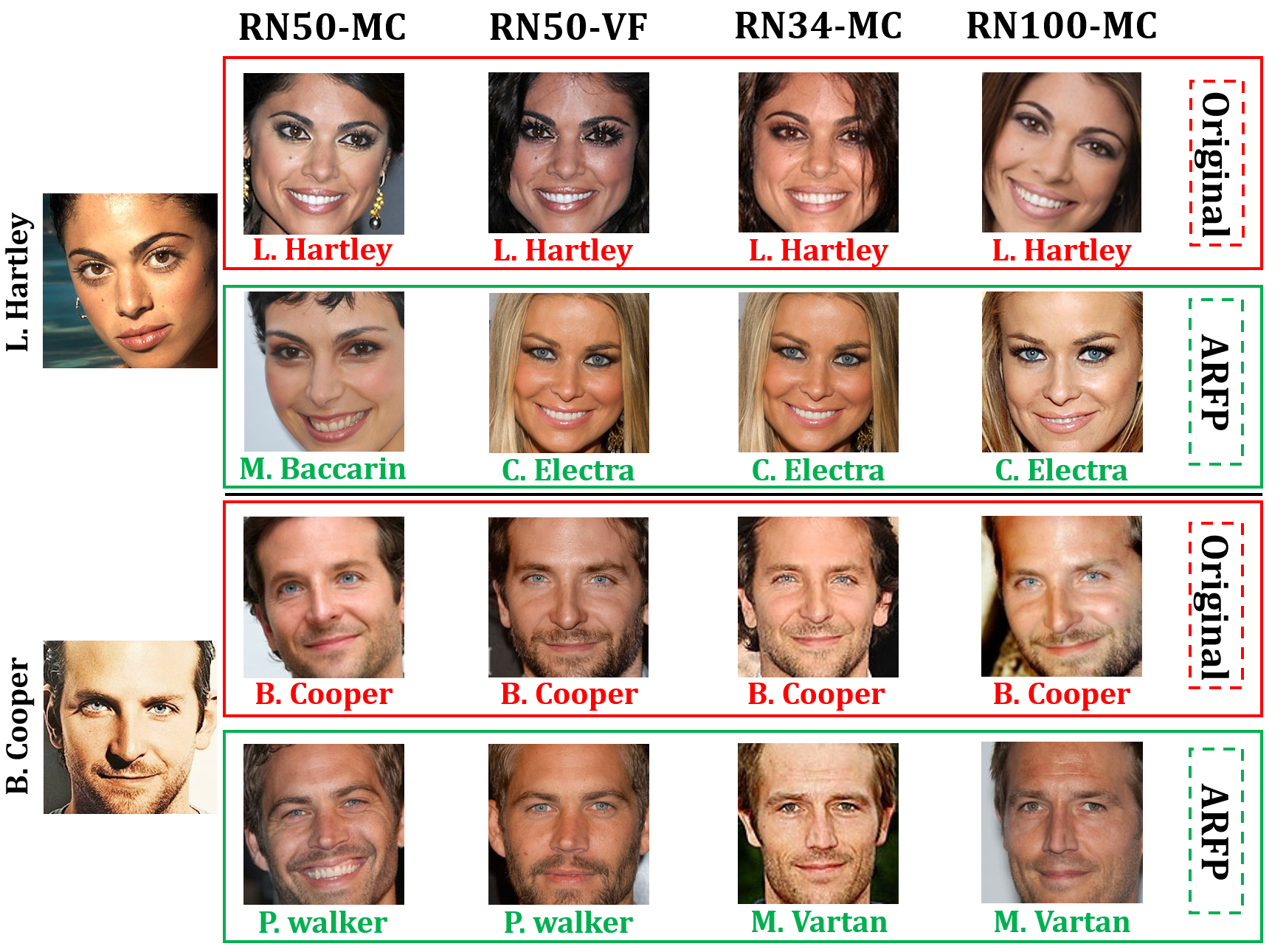}
    \Description{Diagram.}
    \caption{Cross-model retrieval on FaceScrub after protection.}
    \vspace{-30pt}
    \label{fig:facescrub_retrieval}
\end{wrapfigure}
 
To evaluate protection effectiveness and generalization across unseen evaluators, we compare ARFP with Chameleon~\cite{chow2024chameleon}, OPOM~\cite{zhong2022opom}, and TIP-IM~\cite{Yang_2021_ICCV}. As shown in Fig.~\ref{fig:comparison_chart}, ARFP achieves high unseen-model protection, with average PSR above $90\%$, and remains competitive with or better than the reported baselines across standard architectures.

Table~\ref{tab:psr_detailed} provides a finer-grained evaluation by reporting PSR for ten randomly selected identities against eight black-box FR models. These evaluators vary in architecture and pretraining source. ARFP achieves a mean PSR of $95.91\%$, with moderate variance across identities, indicating consistent protection performance across the tested models and subjects.

\begin{wrapfigure}{R}{0.48\textwidth}
    \centering
    \vspace{-20pt}
    \includegraphics[width=0.45\textwidth]{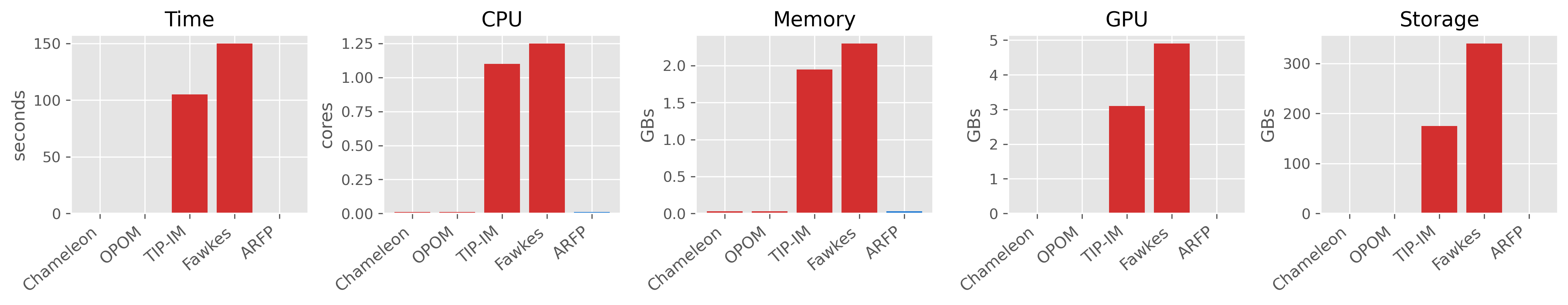}
    \Description{Diagram.}
    \vspace{-10pt}
    \caption{Protection efficiency analysis. ARFP uses a single forward pass, whereas optimization-based baselines incur a higher generation cost.}
    \label{fig:cost_analysis}
    \vspace{-15pt}
\end{wrapfigure}
To qualitatively illustrate black-box transferability, we visualize retrieval results for unauthorized queries. As shown in Fig.~\ref{fig:facescrub_retrieval}, after applying ARFP, the retrieved identities are no longer the true matches.

We also evaluate computational cost, as shown in Fig.~\ref{fig:cost_analysis}. Optimization-based approaches incur substantial overhead during protection generation. In contrast, ARFP requires only a single forward pass, achieving near real-time performance ($\approx 0.05$s) with low CPU and memory overhead, which is favorable for practical deployment.

\begin{table*}[t]
\centering
\caption{Protection Success Rate (PSR \%) of ARFP across eight black-box models.}
\label{tab:psr_detailed}
\resizebox{\textwidth}{!}{
\begin{tabular}{lcccccccc|cc}
\toprule
\textbf{Identity} & \multicolumn{8}{c|}{\textbf{PSR (\%)}} & \multicolumn{2}{c}{\textbf{Statistics}} \\ \cmidrule(r){2-9} \cmidrule(l){10-11}
 & \textbf{EN-MC} & \textbf{RN50-GL} & \textbf{RN50-MC} & \textbf{RN50-VF} & \textbf{RN50-WF} & \textbf{RN18-MC} & \textbf{RN34-MC} & \textbf{RN100-MC} & \textbf{Mean} & \textbf{Std} \\
\midrule
A. Felix & 100.00 & 100.00 & 93.33 & 86.67 & 100.00 & 93.33 & 86.67 & 100.00 & 95.00 & 5.27 \\
B. Gordon & 100.00 & 100.00 & 91.67 & 100.00 & 100.00 & 91.67 & 100.00 & 91.67 & 96.88 & 4.14 \\
C. Myers & 100.00 & 100.00 & 81.82 & 100.00 & 100.00 & 90.91 & 81.82 & 100.00 & 94.32 & 7.91 \\
D. Osorno & 100.00 & 92.31 & 100.00 & 92.31 & 100.00 & 92.31 & 84.62 & 100.00 & 95.19 & 5.56 \\
G. Ferreia & 100.00 & 100.00 & 100.00 & 92.86 & 100.00 & 92.86 & 92.86 & 100.00 & 97.32 & 3.57 \\
I. Perlman & 100.00 & 100.00 & 80.00 & 100.00 & 100.00 & 90.00 & 80.00 & 100.00 & 93.75 & 8.84 \\
M. Georges & 100.00 & 100.00 & 93.33 & 100.00 & 100.00 & 93.33 & 86.67 & 100.00 & 96.67 & 4.71 \\
O. Neuville & 91.67 & 100.00 & 100.00 & 100.00 & 91.67 & 100.00 & 91.67 & 100.00 & 96.88 & 4.14 \\
R. Corrie & 100.00 & 100.00 & 93.75 & 100.00 & 100.00 & 93.75 & 93.75 & 100.00 & 97.66 & 3.12 \\
V. Zhao & 100.00 & 100.00 & 90.91 & 100.00 & 100.00 & 90.91 & 81.82 & 100.00 & 95.45 & 6.57 \\
\midrule
\textbf{Mean} & \textbf{99.17} & \textbf{99.23} & \textbf{92.48} & \textbf{97.18} & \textbf{99.17} & \textbf{92.91} & \textbf{87.99} & \textbf{99.17} & \textbf{95.91} & \textbf{4.22} \\
\bottomrule
\end{tabular}
}
\vspace{-15pt}
\end{table*}

\begin{wrapfigure}{r}{0.48\textwidth}
\vspace{-50pt}
    \includegraphics[width=0.45\textwidth]{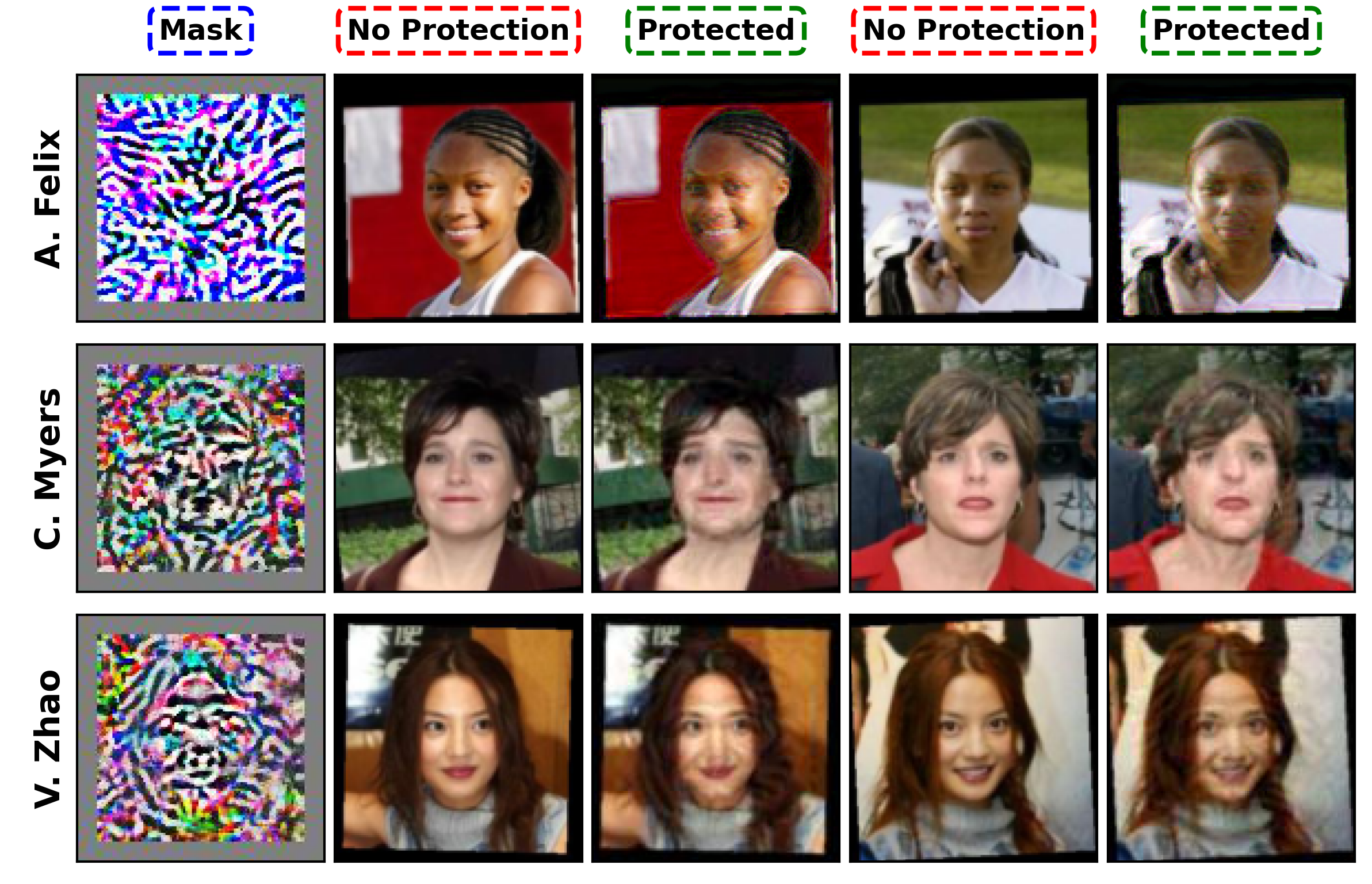}
    \Description{Diagram.}
    \caption{Visual consistency of ARFP across pose and lighting variations.}
    \label{fig:image_quality_consistency}
    \vspace{-40pt}
\end{wrapfigure}

\subsection{Image Quality and Consistency}
Preserving visual quality is an important objective for practical deployment. ARFP is designed to learn a transformation that maintains perceptual consistency while suppressing identity information. As illustrated in Fig.~\ref{fig:image_quality_consistency}, the generated images retain natural appearance and exhibit stable visual characteristics across variations in pose and lighting.

\begin{table}[t]
    \centering
    \caption{Quantitative image quality comparison.}
    \label{tab:quality}
    \resizebox{0.45\textwidth}{!}{
    \begin{tabular}{lcc}
    \toprule
    \textbf{Method} & \textbf{PSNR (dB) $\uparrow$} & \textbf{SSIM $\uparrow$} \\
    \midrule
    Fawkes~\cite{shan2020fawkes} & 22.4 & 0.81 \\
    Chameleon~\cite{chow2024chameleon} & 28.5 & 0.89 \\
    \textbf{ARFP (Ours)} & \textbf{31.2} & \textbf{0.94} \\
    \bottomrule
    \end{tabular}
    }
    \vspace{-10pt}
\end{table}

\begin{wraptable}{r}{0.60\textwidth}
    \vspace{-10pt}
    \caption{Full-scale restoration attack evaluation on the LFW test set. Mean PSR (\%) $\pm$ variance.}
    \label{tab:restoration_full}
    \centering
    \small
    \setlength{\tabcolsep}{4pt}
    \begin{tabular}{lccc}
    \toprule
    \textbf{Method} & \textbf{DiffPure} & \textbf{CycleGAN} & \textbf{BINN} \\
    \midrule
    Fawkes & $23.1 \pm 5.1$ & $18.4 \pm 4.2$ & $10.1 \pm 3.2$ \\
    Chameleon & $70.0 \pm 8.0$ & $32.0 \pm 12.0$ & $22.0 \pm 8.0$ \\
    \textbf{ARFP (Ours)} & $\mathbf{93.4} \pm 1.8$ & $\mathbf{86.5} \pm 2.5$ & $\mathbf{91.2} \pm 2.1$ \\
    \bottomrule
    \end{tabular}
    \vspace{-30pt}
\end{wraptable}

Table~\ref{tab:quality} shows that ARFP also preserves image fidelity well, with a PSNR of 31.2\,dB and an SSIM of 0.94. We next evaluate its robustness against restoration-oriented inversion attacks.

\subsubsection{Robustness Experiment}
We consider three restoration baselines: DiffPure \cite{nie2022diffusion}, CycleGAN \cite{zhu2017unpaired}, and BINN \cite{jacobsen2018revnet}. Table~\ref{tab:restoration_full} summarizes the performance under these attacks. ARFP maintains a mean PSR of $91.2\%$ under BINN, the strongest paired attack in this evaluation. For the evaluated restoration families, these results indicate that the key-conditioned transformation makes recovery of identity-consistent clean images more difficult than for the compared baselines. Visualizations of adversarially attacked images are provided in the Appendix~\ref{sec:appendix_qua}.

\subsection{Ablation Study}

\begin{wraptable}{l}{0.78\textwidth}
    \vspace{-15pt}
    \caption{Ablation of ARFP components. We report PSR before and after the Adaptive Reversal attack, together with SSIM.}
    \label{tab:ablation}
    \begin{tabular}{cccccc}
    \toprule
    \textbf{ARAT} & \textbf{KMB} & \textbf{ARR} & \textbf{Clean PSR (\%)} & \textbf{Attacked PSR (\%)} & \textbf{SSIM} \\
    \midrule
    \texttimes & \checkmark & \checkmark & 93.8 & \textbf{24.5} & \textbf{0.98} \\
    \checkmark & \texttimes & \checkmark & 94.1 & 41.2 & 0.85 \\
    \checkmark & \checkmark & \texttimes & \textbf{96.5} & 88.4 & 0.94 \\
    \checkmark & \checkmark & \checkmark & 95.9 & \textbf{91.2} & 0.94 \\
    \bottomrule
    \end{tabular}
\end{wraptable}

To evaluate the contribution of each module, we conduct the ablation study reported in Table~\ref{tab:ablation}. Here, Clean PSR denotes the defense success rate measured directly on the protected images. Attacked PSR denotes the protection success rate after the protected images are subjected to a BINN attack designed to remove the privacy perturbations. These two metrics quantify robustness under standard and adaptive evaluation settings.

\textbf{The Effect of ARAT.}  
Removing the surrogate restoration adversary leads to a large drop under adaptive reversal. While the model still achieves high Clean PSR (93.8\%), Attacked PSR drops to 24.5\%. This shows that restoration-aware training is important for robustness to the evaluated inversion attack.

\textbf{The Effect of KMB.}  
Without key conditioning, the model becomes a deterministic transformation that is easier to approximate from paired data. As shown in Table~\ref{tab:ablation}, removing the key reduces Attacked PSR to 41.2\% and lowers reconstruction quality to SSIM 0.85, indicating that key-dependent modulation contributes substantially to restoration resistance.

\textbf{The Effect of ARR.}  
Removing the nonce branch slightly improves Clean PSR (96.5\%) but reduces Attacked PSR under purification (88.4\% versus 91.2\%). This suggests that the auxiliary nonce regularization stabilizes the learned transformation and modestly improves robustness in the evaluated setting.

\subsection{Access Control}
\noindent\textbf{Key Sensitivity.}  
As shown in Table~\ref{tab:key_access}, correct-key access enables faithful recovery with high structural consistency (SSIM 0.94), whereas incorrect-key access leads to severe reconstruction failure (SSIM $< 0.4$). Meanwhile, the nonce branch rejects wrong-key outputs with a Bit Error Rate (BER) close to that of random guessing. These results show strong empirical key sensitivity for authorized recovery under the evaluated conditions. In addition, ARFP supports nonce-based integrity-aware recovery as a lightweight tamper-indication mechanism during restoration.

\begin{table}[t]
    \caption{Access control and key sensitivity analysis, assessing the separation between correct-key and wrong-key recovery.}
    \label{tab:key_access}
    \centering
\begin{tabular}{lcccc}
\toprule
\textbf{Access Condition} & \textbf{Key Error (HD)} & \textbf{PSNR $\uparrow$} & \textbf{SSIM $\uparrow$} & \textbf{Nonce BER $\downarrow$} \\
\midrule
\textbf{Authorized} ($k_{true}$) & 0 bits & \textbf{31.2} & \textbf{0.94} & \textbf{1.2\%} \\
\midrule
\textbf{Minor Mistake} & 1 bit & 14.3 & 0.38 & 48.9\% \\
\textbf{Moderate Mistake}& 16 bits & 12.8 & 0.33 & 50.1\% \\
\textbf{Random Guess} & $\approx 128$ bits & 12.1 & 0.31 & 50.4\% \\
\bottomrule
\end{tabular}
\end{table}

\begin{wraptable}{r}{0.4\textwidth}
    \centering
    \caption{BER under different conditions.}
    \label{tab:fingerprint}
    \begin{tabular}{lcc}
    \toprule
    \textbf{Condition} & \textbf{BER (\%)}\\
    \midrule
    Raw & 1.2\% \\
    JPEG & 4.5\% \\
    \midrule
    Splicing~\cite{dong2013casia} & 48.3\% \\
    BINN~\cite{jacobsen2018revnet} & 49.1\% \\
    DiffPure~\cite{nie2022diffusion} & 50.2\% \\
    \bottomrule
    \end{tabular}%
    \vspace{-10pt}
\end{wraptable}
We further evaluate the robustness of integrity-aware recovery under benign and malicious conditions. Benign settings include raw and JPEG-compressed images. Malicious scenarios include local tampering through splicing \cite{dong2013casia}, as well as two restoration-oriented attacks: BINN \cite{jacobsen2018revnet} and DiffPure \cite{nie2022diffusion}. As shown in Table~\ref{tab:fingerprint}, BER remains low under benign conditions and rises sharply under tampered or adversarially transformed inputs. In the evaluated setting, this behavior supports the nonce branch as a practical indicator of tampering or restoration failure, rather than as a general verification mechanism.

\subsection{Hyperparameter Selection}
\begin{wrapfigure}{L}{0.45\textwidth}
    \centering
    \includegraphics[width=0.45\textwidth]{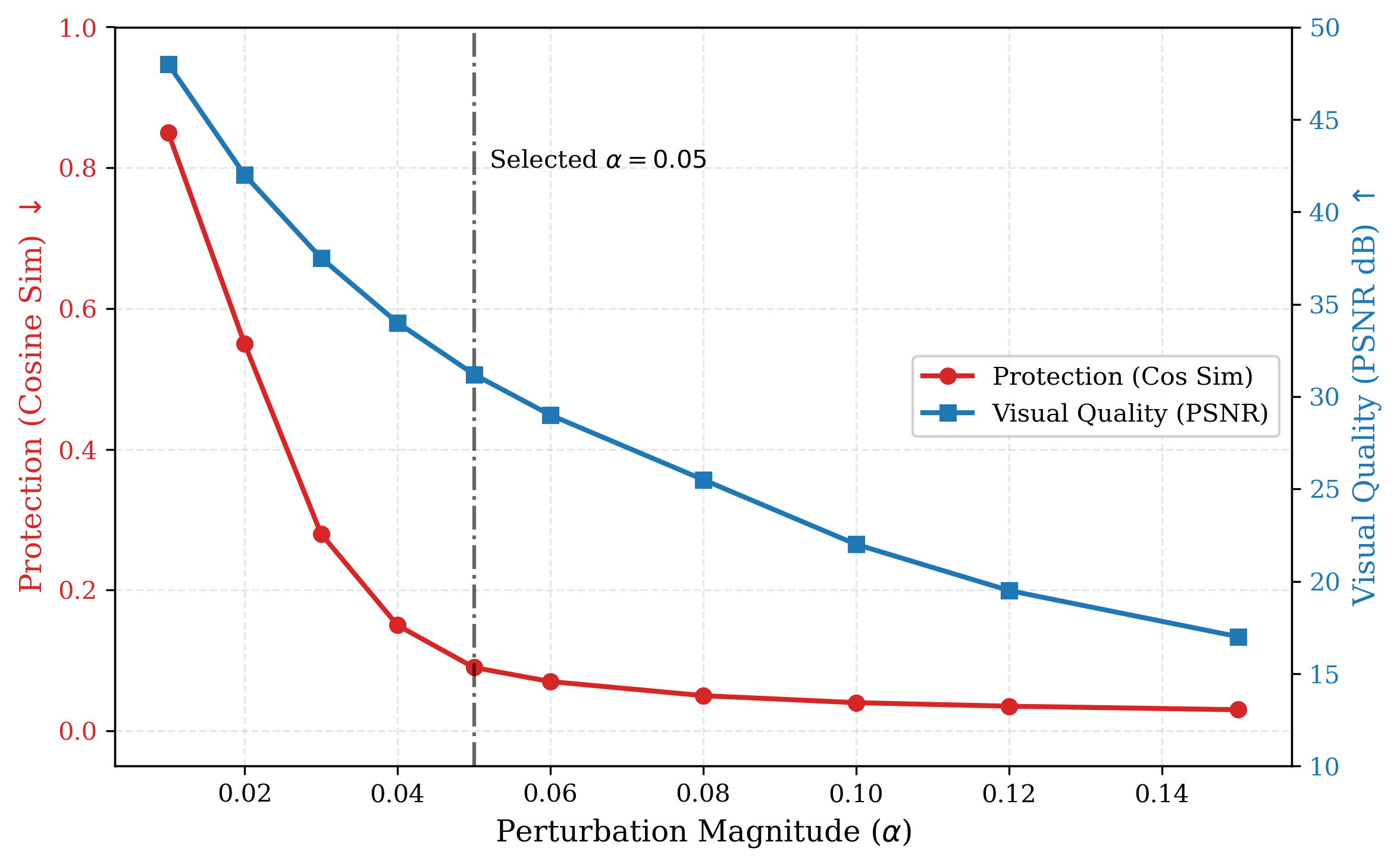}
    \Description{Diagram.}
    \vspace{-10pt}
    \caption{The impact of $\alpha$.}
    \label{fig:tradeoff}
\end{wrapfigure}

We further analyze the sensitivity of ARFP to the perturbation magnitude factor $\alpha$, which controls the trade-off between protection strength and visual quality. As illustrated in Figure~\ref{fig:tradeoff}, varying $\alpha$ from 0.01 to 0.15 reveals three regimes. For small perturbations ($\alpha < 0.03$), visual quality remains high (PSNR $> 35$ dB), but protection against strong FR models is insufficient. Around $\alpha = 0.05$, the curve exhibits a knee point: further increases produce smaller gains in protection while degrading perceptual quality more noticeably. We therefore select $\alpha = 0.05$ as the operating point used in the main experiments. For larger perturbations ($\alpha > 0.08$), protection continues to increase, but visible artifacts become more pronounced and PSNR declines.

\section{Conclusion}

In this paper, we identify the lack of controllable reversibility in existing face privacy protection methods, which exposes them to asymmetric adversarial attacks. To address this, we introduce an attack paradigm and propose the ARFP framework, formulating face privacy as an asymmetric bidirectional problem. ARFP integrates key-conditioned generation, adversarial training against inversion, and authorized recovery, enabling robust protection against adaptive attacks while preserving reliable recovery. In addition, a nonce-based mechanism provides a lightweight signal for tamper awareness. Extensive experiments demonstrate that ARFP achieves strong robustness, high visual fidelity, and efficient deployment. Future work includes extending the framework to broader modalities and exploring stronger theoretical guarantees for controllable reversibility.

\nocite{*}
\bibliographystyle{plainnat}
\bibliography{references}

\newpage
\appendix

\section{Appendix: Supporting Derivation for Proposition~\ref{prop:feature_leakage}}
\label{sec:appendix_proof}

This appendix provides a short derivation supporting Proposition~\ref{prop:feature_leakage}. The goal is not to establish a formal security theorem, but to clarify why reduced mutual information between the protected output and the identity embedding is consistent with increased difficulty of feature estimation.

Let
\[
h = \phi(x) \in \mathbb{R}^{d}
\]
denote the identity embedding extracted from the clean input \(x\), and let
\[
z = G(x,k)
\]
denote the protected output generated from \(x\) and key \(k\). We consider an adversary that estimates \(h\) from \(z\) under squared error.

For any estimator \(A\), the expected reconstruction error is
\begin{equation}
\mathbb{E}\!\left[\|A(z)-h\|_{2}^{2}\right].
\end{equation}
The estimator that minimizes this quantity is the conditional expectation
\begin{equation}
A^{*}(z) = \mathbb{E}[h \mid z],
\end{equation}
and the corresponding minimum mean-squared error is
\begin{equation}
\operatorname{MMSE}(h \mid z)
=
\inf_{A}\mathbb{E}\!\left[\|A(z)-h\|_{2}^{2}\right]
=
\mathbb{E}\!\left[\|h-\mathbb{E}[h \mid z]\|_{2}^{2}\right].
\end{equation}
This identity shows that the best achievable feature-estimation performance is determined by the conditional uncertainty of \(h\) given the protected observation \(z\).

From an information-theoretic perspective, the uncertainty in \(h\) after observing \(z\) is characterized by the conditional entropy \(H(h \mid z)\) or, in the continuous setting, by the conditional differential entropy \(h(h \mid z)\). Moreover, the mutual information decomposes the uncertainty in \(h\) into a marginal term and a conditional term. Therefore, if the mutual information between \(h\) and \(z\) is reduced while the marginal uncertainty of \(h\) remains fixed, the uncertainty that remains after observing \(z\) becomes larger.

Under standard regularity conditions, larger conditional uncertainty is generally associated with weaker recoverability of the identity embedding from the protected observation. This is the sense in which the protection mechanism is intended to reduce feature leakage: by decreasing the information about \(h\) that is preserved in \(z\), it makes identity-related feature estimation more difficult for an adversary on average.

We emphasize that this derivation is interpretive rather than a proof of a universal closed-form lower bound on estimation error solely in terms of mutual information. It does not establish cryptographic irreversibility, formal inversion resistance, or provable security. Its role is limited to providing a principled explanation for why reducing feature leakage is a meaningful objective for the proposed key-conditioned protection mechanism.

\section{Experimental Setup Details}
\label{sec:appendix_exp}

\subsection{Datasets and Protocols}
(1) LFW: Represents an unconstrained recognition scenario with variations in pose, lighting, and expression. Following the protocol in~\cite{chow2024chameleon,cherepanova2021lowkey}, we filter identities with at least 15 images to simulate users who actively post photos online. For each identity, we randomly split images into a reference set (for protection generation) and a probe set (for evaluation), ensuring no data leakage.

(2) FaceScrub: Used for identity-level robustness and generalization analysis. Its broad identity coverage supports rigorous evaluation of protection transferability across subjects.

All facial images in both datasets are pre-processed using MTCNN~\cite{zhang2016joint} for detection and alignment, then resized to $112 \times 112$ resolution, and finally normalized to the $[-1, 1]$ standard range before being fed into the protection network.

\subsection{Face Recognition Models}
To comprehensively evaluate the robustness and transferability of our protection, we employ the same diverse collection of publicly available pretrained FR models as in Chameleon~\cite{chow2024chameleon}, detailed in Table~\ref{tab:fr_models}. These models cover a wide range of architectures and training subsets.

\begin{table}[h]
    \centering
    \caption{The collection of publicly available pretrained FR models.}
    \label{tab:fr_models}
    \begin{tabular}{cccc}
    \toprule
    \textbf{Model ID} & \textbf{Neural Arch.} & \textbf{Training Dataset} & \textbf{FR Acc.} \\
    \midrule
    EN-MC & EfficientNet & MS-Celeb-1M & 94.94\% \\
    RN50-GL & ResNet50 & Glint360K & 96.68\% \\
    RN50-MC & ResNet50 & MS-Celeb-1M & 89.25\% \\
    RN50-VF & ResNet50 & VGG-Face2 & 94.30\% \\
    RN50-WF & ResNet50 & WebFace600K & 96.72\% \\
    RN18-MC & ResNet18 & MS-Celeb-1M & 82.64\% \\
    RN34-MC & ResNet34 & MS-Celeb-1M & 84.49\% \\
    RN100-MC & ResNet100 & MS-Celeb-1M & 91.85\% \\
    \bottomrule
    \end{tabular}
\end{table}

\subsection{Implementation Details}
All experiments are implemented in TensorFlow and executed on a single NVIDIA RTX 4060 GPU. The protection generator follows the ResNet-based key-conditioned architecture described in the main method section, and the condition encoder is implemented as a lightweight MLP. Following the optimization protocol in Chameleon~\cite{chow2024chameleon}, we use the Adam optimizer with a learning rate of $2 \times 10^{-4}$ and a batch size of 16. The balancing hyperparameters are set to $\lambda_{rec}=10$, $\lambda_{id}=5$, and $\lambda_{rev}=1.0$, which were found empirically to provide stable convergence.

To model adaptive restoration attacks, we train a separate reversal network on the clean LFW training split until convergence and use it as the surrogate adversary in evaluation. The secret key $k$ and nonce $m$ are encoded as 256-bit and 64-bit binary vectors, respectively. The identity loss $\mathcal{L}_{id}$ is implemented as a cosine embedding loss over intermediate features extracted by a pretrained FR network. For paired inverse-attacker training, the surrogate adversary uses a batch size of 32, an initial learning rate of $10^{-4}$ decayed by 0.95 every 10 epochs, and paired samples generated from the underlying dataset.\footnote{The code will be made publicly available after review; it is omitted here for anonymity.}

\section{Qualitative purification examples}
\label{sec:appendix_qua}
\begin{wrapfigure}{R}{0.48\textwidth}
\vspace{-20pt}
    \includegraphics[width=0.45\textwidth]{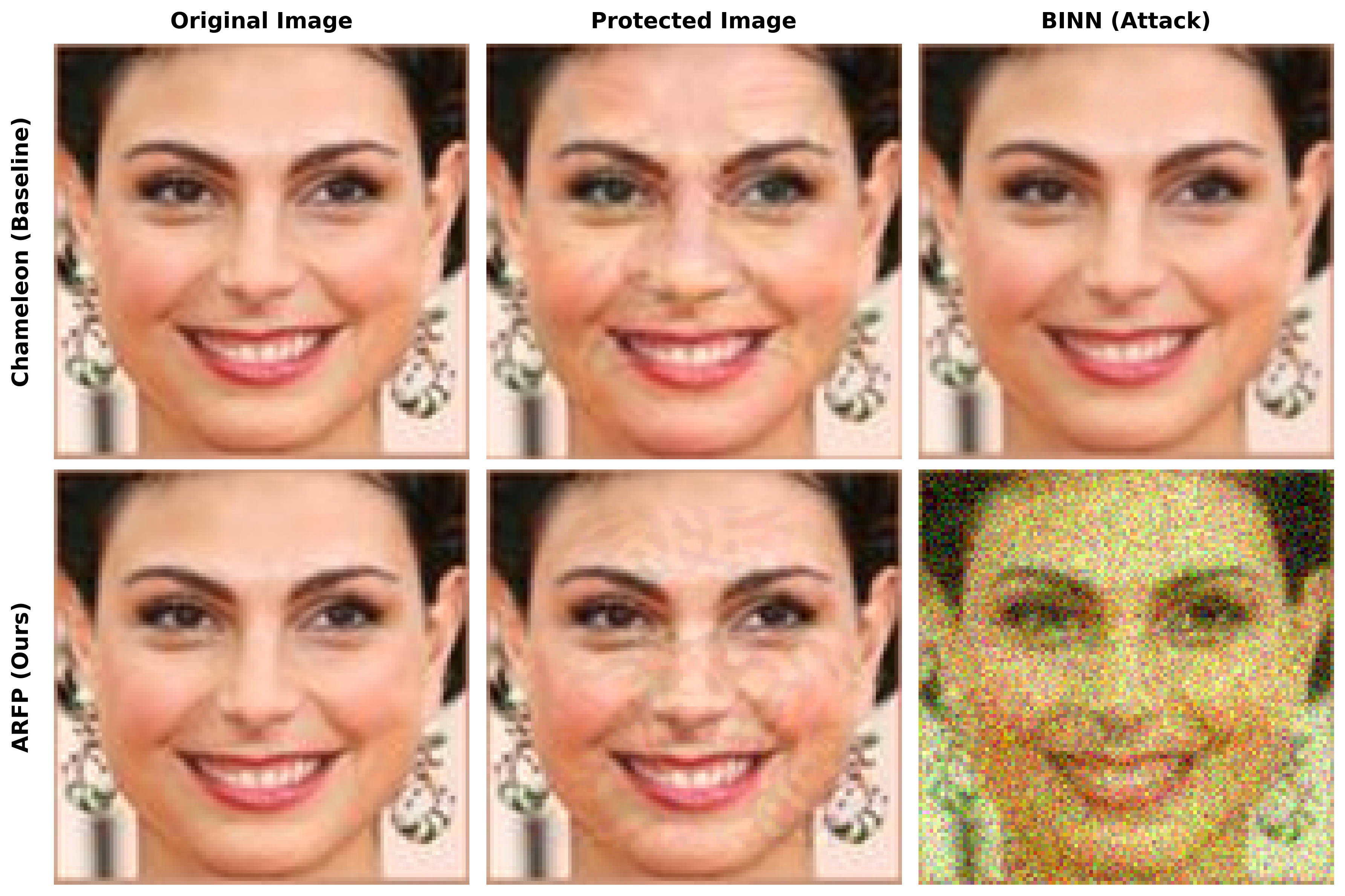}
    \Description{Diagram.}
    \caption{Qualitative purification examples. The evaluated attack procedure recovers more identity-consistent structure from Chameleon outputs (Top), whereas ARFP outputs (Bottom) remain substantially distorted under the same procedure.}
    \label{fig:purification_viz}
    \vspace{-40pt}
\end{wrapfigure}

As shown in Fig.~\ref{fig:purification_viz}, the evaluated purification procedure produces visually more identity-consistent outputs for Chameleon than for ARFP. Under the same attack setting, the ARFP examples remain heavily distorted, indicating weaker recovery quality for the attacker in this qualitative comparison. These examples should be interpreted as illustrative evidence for the evaluated procedure rather than as a general statement about all possible restoration attacks.

\section{Discussion and Limitations}
While ARFP demonstrates strong protection capabilities, there are several aspects worth discussing regarding its practical deployment.

Computational Overhead.
Compared to traditional additive noise attacks like Fawkes, our method involves a forward pass through an adaptive reversal network, which is slightly more computationally intensive but still efficient. The primary cost lies in the training phase, where the adversarial loop requires oscillating updates. However once trained, the generation of personalized protection is highly efficient due to its feed-forward design, making it feasible for real-time user-end applications.

Physical World Robustness.
Our current evaluation focuses on the digital domain. In real-world scenarios where protected images might be printed and recaptured (print-scan attack), the high-frequency components of the adversarial perturbation and the fragile integrity indicator (Nonce) might be degraded. Enhancing the robustness of both the protection and the watermark against analog transformations remains a direction for future work.

Key Management.
The security of our reversibility relies entirely on the secrecy of the user key $k_u$. This design choice, while structurally sound, introduces a key management burden. If a user loses their key, the original image cannot be mathematically recovered from the protected version. This intentional "failure" ensures that no backdoor exists for unauthorized parties, but it requires users to securely store their keys.

\end{document}